\documentclass[a4paper, 10pt, twocolumn, twoside]{article}

\usepackage{SASBEFRBDTSC}

\usepackage{lscape}
\usepackage{hologo}

\usepackage{algorithm}
\usepackage{algorithmic}
\usepackage{amsmath}
\usepackage{subcaption}
\usepackage{newfloat}
\usepackage{listings}

\begin{document}

\linespread{0.5}

\title{Bi-level RL-Heuristic Optimization for Real-world Winter Road Maintenance}

\author{Yue Xie$^{1}$\thanks{This project has received funding from the European Union’s Horizon 2020 research and innovation programme under the Marie Skłodowska-Curie grant agreement No 101034337.}, and  Zizhen Xu$^{2}$ and William Beazley$^{3}$ and Fumiya Iida $^{2}$}

\affiliation{
$^1$School of Science, Loughborough University, UK\\
$^2$Department of Engineering, University of Cambridge, UK \\
$^3$National Highways, 199 Wharfside St, Birmingham B1 1RN, UK
}

\email{
\href{mailto:Y.Xie4@lboro.ac.uk}{Y.Xie4@lboro.ac.uk}, 
\href{mailto:zx311@cam.ac.uk}{zx311@cam.ac.uk}
}

\maketitle 
\thispagestyle{fancy} 
\pagestyle{fancy}

\begin{abstract}
Winter road maintenance is critical for ensuring public safety and reducing environmental impacts, yet existing methods struggle to manage large-scale routing problems effectively and mostly reply on human decision. This study presents a novel, scalable bi-level optimization framework, validated on real operational data on UK strategic road networks (M25, M6, A1), including interconnected local road networks in surrounding areas for vehicle traversing, as part of the highway operator's efforts to solve existing planning challenges. At the upper level, a reinforcement learning (RL) agent strategically partitions the road network into manageable clusters and optimally allocates resources from multiple depots. At the lower level, a multi-objective vehicle routing problem (VRP) is solved within each cluster, minimizing the maximum vehicle travel time and total carbon emissions. Unlike existing approaches, our method handles large-scale, real-world networks efficiently, explicitly incorporating vehicle-specific constraints, depot capacities, and road segment requirements. Results demonstrate significant improvements, including balanced workloads, reduced maximum travel times below the targeted two-hour threshold, lower emissions, and substantial cost savings. This study illustrates how advanced AI-driven bi-level optimization can directly enhance operational decision-making in real-world transportation and logistics.
\end{abstract}


\section{Introduction}

Ensuring efficient winter road maintenance is essential for safeguarding public safety, sustaining economic activity, and mitigating environmental impact-particularly in regions subject to severe weather. Effective routing of salt-spreading vehicles is a cornerstone of winter road management, directly influencing road safety and indirectly affecting operational costs and carbon emissions. However, real-world winter maintenance remains challenging due to the scale and complexity of motorway networks, the presence of multiple depots with varying capacities, heterogeneous vehicle fleets, and stringent operational constraints. Traditional approaches such as manual or heuristic are struggle to scale, resulting in suboptimal resource utilization, excessive emissions, and delayed responses during critical weather events.

Local highway authorities—such as Liverpool City Council, South Lanarkshire County Council, Carmarthenshire County Council, and Lincolnshire County Council—have demonstrated that significant financial, operational, and carbon savings can be achieved through the effective optimization of winter treatment routes, while maintaining a safe and resilient road network during hazardous conditions. These successes, however, have largely been realized on smaller, local networks. Strategic road networks differ substantially in both scale and operational nature, meaning that approaches proven at the local level often struggle—or fail outright—to optimize effectively at the national or regional scale.

For strategic network operators, there are two major optimization challenges. The first one is the network-wide coordination, ensuring that regional and overall network savings are fully realized, including the routing of large multi-line motorways (four or more lanes) and the management of complex junctions. The second challenge is the constraint-aware routing, where operational requirements may mandate secondary vehicle passes or multiple treatments over the same road segment, requiring more detailed modeling and optimization. 

The first challenge stems from the sheer scale and complexity of junctions, depot placements, and interconnections between strategic and local roads, often exceeding the capacity of traditional optimization techniques without substantially more computational power or novel algorithmic approaches. The second challenge demands fine-grained operational modeling, as overlapping treatments and route dependencies introduce combinatorial complexity. Beyond the immediate operational and environmental benefits, effective optimization of these large-scale networks could also serve as a foundation for future innovations in winter maintenance, including advanced route digitization, dynamic spreading strategies, and automated treatment verification.

In response, we introduce a bi-level optimization framework that combines reinforcement learning for strategic decision-making with constraint-aware vehicle routing optimization. The upper level uses a Proximal Policy Optimization (PPO) agent to generate depot and vehicle assignments, initially guided by a KDTree-based spatial clustering for efficient network partitioning~\cite{bentley1975multidimensional,friedman1977algorithm}. The lower level then solves a multi-objective vehicle routing problem (VRP) within each assigned cluster, producing feasible, high-quality routes that explicitly optimize for both the travel time and environmental impact.

We validate our approach using realistic operational data from prominent UK strategic road networks, including sections of the M25, M6, and A1. Experimental results demonstrate significant improvements over existing heuristic methods, achieving a substantial reduction in maximum route completion times (ensuring all vehicles complete their tasks within a targeted two-hour time window), lowering total carbon emissions, and enhancing the equitable distribution of workloads across depots and vehicles. These results substantiate our claims that intelligent, bi-level optimization approaches can deliver practical and measurable benefits in complex, large-scale operational scenarios, contributing directly to enhanced public safety, environmental responsibility, and operational efficiency in winter road maintenance.

\section{Related Work}
\label{sec:related}

Winter service operations, including gritting and de-icing, are often formulated as arc routing problems, where the objective is to service road segments rather than visit discrete customers. The Capacitated Arc Routing Problem (CARP)~\cite{golden1981capacitated,corberan2015arc} and its extensions have been extensively studied, with heuristic and metaheuristic approaches such as tabu search~\cite{hertz2000tabu}, memetic algorithms~\cite{mei2011memetic}, and hybrid genetic algorithms~\cite{jozefowiez2009evolutionary,sitek2021optimization} achieving strong results for small- and medium-scale instances. For winter maintenance specifically, variants such as the Winter Road Maintenance Problem (WRMP)~\cite{perrier2006survey,perrier2006survey1} has incorporated domain-specific constraints including lane-dependent service, spreader capacity, time windows, and mandatory return-to-depot policies. These classical approaches have proven effective at a municipal scale but struggle to scale computationally when applied to strategic road networks containing thousands of segments, multiple depots, and intricate operational requirements, as well as interconnected local road networks that are usually used for vehicle traversing. 

In the UK, operational guidelines and performance targets for winter service are documented in the \emph{Well-managed Highway Infrastructure: A Code of Practice}\cite{uk2016well,councilwinter} and annual reports by the Department for Transport\cite{adams2003performance, walsh2016winter}. Local authority reviews, such as the \emph{Winter Service Policy and Plan} from Liverpool City Council~\cite{Sefton_Winter_Service_Policy_v4}, South Lanarkshire Council’s \emph{Winter Maintenance Plan}\cite{SouthLanarkshire_Winter_Service_2024_2025}, and Carmarthenshire County Council’s winter service performance reports\cite{Carmarthenshire_Winter_Service_Plan}, demonstrate measurable financial and environmental benefits from optimized winter treatment routes. However, these studies primarily address smaller, district-level networks, with methods that often do not generalize to high-capacity motorway systems. The UK National Highways’ \emph{Severe Weather Service Review}~\cite{NationalHighwaysARP4_2024} further highlights scalability challenges in multi-depot motorway coverage and the operational complexity of treating multi-lane carriageways and complex junctions under strict time constraints.

In the context of Operation research, machine learning for combinatorial optimization has explored learned constructive and improvement policies for routing and scheduling. Attention-based neural solvers and reinforcement learning (RL) methods learn to build or refine tours for TSP/VRP-like problems \cite{kool2018attention,lu2019learning}. Particularly relevant are hybrid, bi-level ideas that use learning to drive high-level structural decisions while leaving low-level search to powerful heuristics \cite{yin2024research,jiang2025surrogate}. \cite{wang2021bi} propose a bi-level RL–heuristic framework in which an upper-level RL agent modifies problem structure and a lower-level heuristic quickly solves the induced instance, mitigating sparse rewards and shrinking the action space. Related work in learning to assist mathematical programming (e.g., learning cuts/branching) further supports the value of ML-guided problem reformulation for scalability \cite{tang2020reinforcement}.

Environmental objectives have become central in winter operations, motivating multi-objective formulations that balance completion time and emissions. Green VRP studies model fuel and CO\textsubscript{2} explicitly \cite{sabet2022green,bektacs2011pollution}, supporting lower-level solvers that optimize multiple criteria under operational constraints.

Our work builds on these strands with a scalable bi-level design: an upper-level RL policy partitions and allocates segments across depots (strategic planning), while a lower-level routing heuristic (tactical execution) solves each cluster subject to capacity, time, and directional constraints. This mirrors operational practice and improves tractability on real motorway networks relative to single-level models.

\section{Problem Definition}
\label{sec:problem}

\begin{table*}[th]
    \centering
        \caption{Mathematical Notation}
          \scalebox{0.8}{
    \begin{tabular}{|l|l|}
    \hline
       \textbf{Sets}  &  \\
       $R_C$   & Set of road segments requiring salt coverage $(R_C = \{1,2,\ldots,N\})$\\
       $R_{NC}$ & Set of road segments not requiring salt coverage, used only for connectivity \\
       & $(R_{NC} = \{1,2,\ldots,M\})$ \\
     $R = R_C + R_{NC}$ & Set of road segments\\
       D & Set of depots (each depot associated with vehicle availability constraints) \\
       V & Set of vehicles available for routing $(V=\{1,2,\ldots, K\})$\\
       \hline
       \textbf{Parameters} & \\
       $l_i$ & Length of road segment $i(km)$ \\
       $\delta_i$ & Binary indicator (1 if road is one-way, 0 if two-way)\\
       $v_i^{max}$ & Maximum allowable speed on road segment $i (km/h)$, sourced from OpenStreetMap\\
       $n_i$ & Number of lanes on road segment $i$ \\
       $v^{op}_k$ & Operational (spreading) speed of vehicle $k (km/h)$ \\
       $Q_k$ & Maximum salt-spreading capacity of vehicle $k (km \cdot lane)$, here $Q_k=166$ assuming one type of salt during operation \\
       $f_k$ & Fuel consumption rate of vehicle $k (liters/km)$ \\
       $e_k$ & Carbon emission factor of vehicle $k (kg Co_2/liter fuel)$, here $e_k=2.51$ based on diesel \cite{webcar} \\
       $c^{op}_k$ & Operational cost of vehicle $k (currency/km)$ \\
       $w_k$ & Weight of vehicle $k (kg)$, influencing fuel economy \\
        $A$ & The adjacency matrix of graph model $G$, \\
    & and $A_{ij} =1$ means that there exists a connection from road segment $i$ to $j$.\\
       \hline
       \textbf{Decision Variables} & \\
       $x_{ijk}$ & Binary variable, 1 if vehicle travel directly from segment $i$ to $j$, otherwise 0\\
       $y_k$ & Binary variable, 1 if vehicle is utilized in the solution, otherwise 0\\
       $z_{ik}$ & Binary variable, 1 if vehicle treat segment $i$, otherwise 0\\
       \hline
    \end{tabular}}
    \label{tab:my_label}
\end{table*}

To clearly define the winter road maintenance problem, Table~\ref{tab:my_label} summarizes all relevant decision variables, parameters, and their descriptions.

The winter road maintenance routing problem is formulated as a multi-objective optimization model as follows:

\begin{align}
     \min \quad & Z_1 = max_{k\in V} \biggl(\sum_{i\in R_C } \frac{l_i}{v^{op}_k}z_{ik} \nonumber \\  & + \sum_{i\in R \cup D} \sum_{j\in R \cup D}  \frac{l_i}{v_i^{max}}(x_{ijk}-z_{ik}\biggr )  \label{obj1} \\
     \min \quad & Z_2 = \sum_{k\in V} \sum_{i\in R \cup D} \sum_{j\in R \cup D} (l_i \cdot f_k \cdot e_k) \cdot x_{ijk} \label{obj2}\\
    s.t.\quad  & z_{ik} \leq \sum_{j \in R\cup D} x_{ijk} \quad \forall i\in R_C, k\in V \label{con:treatment_and_travel} \\
    &     v^{op}_k \leq  v_i^{max}, \quad \forall i \in R, k\in V \label{con:Vehicle_speed}\\
    &     v^{op}_k = min( v_i^{max}, 50), \quad \forall i \in R, k\in V  \label{con:specifically}\\
    &   \sum_{i \in R_C } \sum_{j\in R_C \cup D} (l_i \cdot n_i) z_{ik} \leq Q_k, \quad \forall k\in V \label{con:Vehicle_Capacity}\\
    & \sum_{k \in K} \sum_{j \in R} x_{djk}  \leq Depot\ Capacity, \quad \forall d\in D \label{con:Depot_Capacity} \\
    &     \sum_{j \in R} x_{djk} =y_k, \quad  \sum_{i \in R} x_{idk} =y_k, \quad \forall k\in V, d\in D \label{con:Route_Continuity} \\
    & \sum_{i \in R \cup D} \sum_{j \in R \cup D} l_i \cdot x_{ijk}  \leq Max\ Distance, \quad \forall k\in V  \label{con:Operational_Distance_Limit}\\
    &  x_{ijk} \leq A_{ij}, \quad \forall i,j \in R, k\in V \label{con:Road_Directionality1} \\
    &     x_{ijk} \leq 1- \delta_i, \forall i,j \in R, k\in V \label{con:Road_Directionality_twoway} .
\end{align}

The optimization model comprises two objectives that reflect operational efficiency and environmental sustainability. The first objective~\eqref{obj1} aims to minimize makespan, meaning to minimize the maximum travel time among all maintenance vehicles. This objective ensures balanced workloads and adherence to operational timelines, typically aiming for maintenance activities to conclude within a specific time window (for practical target of the highway operator, two hours). Mathematically, this objective measures both the actual treatment time and travel time spent traversing non-treated segments. The second objective~\eqref{obj2} addresses sustainability concerns by minimizing the total carbon emissions generated by the vehicle fleet. This calculation integrates vehicle-specific fuel consumption rates, segment distances, and fuel emission factors, directly linking environmental impact to the selected routing strategy.

The model integrates several practical constraints that ensure realistic and feasible solutions. Treatment Coverage Constraint~\eqref{con:treatment_and_travel} ensures every segment requiring salt treatment is visited and treated exactly once, and only segments traversed by a maintenance vehicle can be treated. Vehicle Speed Constraints~\eqref{con:Vehicle_speed} and~\eqref{con:specifically} guarantee vehicle operational speeds are limited by the road-specific speed limits and an additional safety-imposed operational maximum speed. Vehicle and Depot Capacity Constraints~\eqref{con:Vehicle_Capacity} and~\eqref{con:Depot_Capacity} limit the salt-spreading capacity of each vehicle based on its design specification and constrains depot utilization according to their respective vehicle availability limits. Route Continuity Constraint~\eqref{con:Route_Continuity} guarantees that vehicle routes must start and finish at the same depot, ensuring continuity and operational feasibility. Operational Distance Constraint~\eqref{con:Operational_Distance_Limit} restricts total travel distance of each vehicle, ensuring routes remain practical, efficient, and within fuel and operational limits. Road Directionality Constraints \eqref{con:Road_Directionality1} and \eqref{con:Road_Directionality_twoway} ensure compliance with real-world road configurations, distinguishing between one-way and two-way road segments and preventing infeasible vehicle maneuvers.

Through this structured mathematical model, our optimization formulation rigorously captures the complexity and real-world constraints of winter road maintenance, facilitating effective decision-making with tangible operational and environmental benefits.

\section{Practical Bi-Level Optimization for Multi-Depot Winter Operations}
\label{sec:approach}

We present a bi-level optimization framework that integrated a data-driven upper-level policy with a constraint-aware lower-level solver to plan winter road treatment at motorway scale. The design illustrates how network operators make decisions in practice, first allocate responsibility for required segments to depots, then construct feasible vehicle tours that respect safety, operational, and regulatory rules. This decomposition keeps the method tractable on large networks while ensuring every produced plan is executable.

\subsection{Bi-level Problem Formulation}



Let $G=(V,E)$ be the directed road graph, $D\subset V$ the depots, and $R_C\subset E$ the set of edges requiring treatment. For a depot assignment $\mathbf{a}\in D^{|R_C|}$ and corresponding feasible routes $\mathbf{x}\in\mathcal{X}(\mathbf{a})$, we minimize the sum of two operator-facing objectives: makespan ($Z_1$, the maximum single-vehicle route time) and total emissions of all routes ($Z_2$):

\begin{align}
& \min_{\mathbf{a}}~F(\mathbf{x}^\star(\mathbf{a}),\mathbf{a})~=~Z_1(\mathbf{x}^\star)+Z_2(\mathbf{x}^\star) \\
& \quad\text{s.t.}\quad
\mathbf{x}^\star(\mathbf{a})\in\arg\min_{\mathbf{x}\in\mathcal{X}(\mathbf{a})}~f(\mathbf{x};\mathbf{a}).
\end{align}

The feasible set $\mathcal{X}(\mathbf{a})$ encodes operational constraints: one-way travel, lane-dependent salt usage, vehicle salt capacity (km$\cdot$lane), maximum route duration (e.g., 120\,min), return-to-depot, speed limits (capped by spreading speed), and maximum distance.

\subsection{Upper level approach}

We cast segment-to-depot assignment as a sequential decision problem. The policy observes normalized per-segment features (coordinates, length, speed limit, lane count, and distances to depots) and outputs a depot index. To start from a safe and interpretable baseline, we warm-star the policy with a deterministic KDTree nearest-depot assignment, yielding spatially coherent clusters and immediate feasibility. We then refine this policy using Proximal Policy Optimization (PPO). After a full assignment is proposed, the lower level solves routes and returns $(Z_1,Z_2)$; the RL reward is $r=-(Z_1+Z_2)$, directly aligned with the operator objectives.

The full training procedure is outlined in Algorithm~\ref{alg:up}. In each episode, the agent proposes a depot assignment, receives feedback from the lower level, computes a reward, and updates the policy accordingly. Over time, this iterative process yields an improved assignment strategy that outperforms static heuristics.

\begin{algorithm}[ht]
\caption{Training Procedure for Upper-Level Depot Assignment Policy}
\label{alg:up}
\begin{algorithmic}[1]
\STATE Initialize depot assignment policy $\pi_\theta$ using KDTree heuristic
\FOR{each training episode}
\STATE Encode state features for all treatment-required segments
\FOR{each segment $e$}
\STATE Select depot assignment $a_e \sim \pi_\theta(\cdot \mid s_e)$
\ENDFOR
\STATE Form assignment vector $\mathbf{a} = {a_e}$ for all segments
\STATE Invoke lower-level routing solver using assignment $\mathbf{a}$
\STATE Receive routing objectives: makespan $Z_1$, emissions $Z_2$
\STATE Compute reward $r = - (w_1 Z_1 + w_2 Z_2)$
\STATE Update policy parameters $\theta$ via PPO using reward $r$
\ENDFOR
\end{algorithmic}
\end{algorithm}

\subsection{Lower Level approach}

Given the clusters induced by the upper level, we solve each depot subproblem with a nearest-neighbor (NN) heuristic (cf. Algorithm \ref{alg:low}) augmented with feasibility checks. Starting at the depot, the solver greedily adds the nearest untreated required edge whose insertion keeps all constraints satisfied; when no feasible insertion remains, the vehicle returns to the depot and a new route starts. This ensures full coverage of $R_C$, strict adherence to capacity/duration/speed/one-way rules, and reproducibility for audit and safety review. Emissions are computed from traveled distance and fuel factors.

\begin{algorithm}[ht]
\caption{Lower-level Routing for Each Depot (Constraint-Aware Heuristic)}
\label{alg:low}
\begin{algorithmic}[1]
\FOR{each depot $d$ in assignment $\mathbf{a}$}
\STATE $U \leftarrow$ set of untreated edges assigned to $d$
\WHILE{$U \neq \emptyset$}
\STATE Start new vehicle route at depot $d$
\STATE Insert farthest untreated edge as initial candidate
\WHILE{insertion of next eligible edge does not violate constraints and $U \neq \emptyset$}
\STATE Select next farthest feasible untreated edge
\STATE Insert into current route, update cumulative constraints
\ENDWHILE
\STATE Close route by returning to depot, mark traversed edges as treated, remove from $U$
\ENDWHILE
\ENDFOR
\STATE Return all depot-wise routes, calculate makespan and emissions
\end{algorithmic}
\end{algorithm}


\subsection{Closed-loop bi-level optimization workflow}

Our system forms a tight feedback loop between assignment and routing. The upper-level PPO policy observes normalized road-segment features (geometry, length, speed, lane count, and distance to depots) and proposes a depot assignment for each required segment. A KDTree-based nearest-depot map provides a warm-start (shown as a dashed initialization), ensuring spatially coherent assignments from the first iteration. Given the assignment, the lower-level solver constructs feasible tours per depot using a constraint-aware NN heuristic that enforces operational rules ( less then 2h per vehicle, less then 630 km travel, and capacity per lane). The routing layer returns objective metrics and policy receives a scalar reward. The loop repeats for several iterations, improving assignments while always producing regulation-compliant routes. 

\begin{figure}
    \centering
    \includegraphics[width=0.95\linewidth]{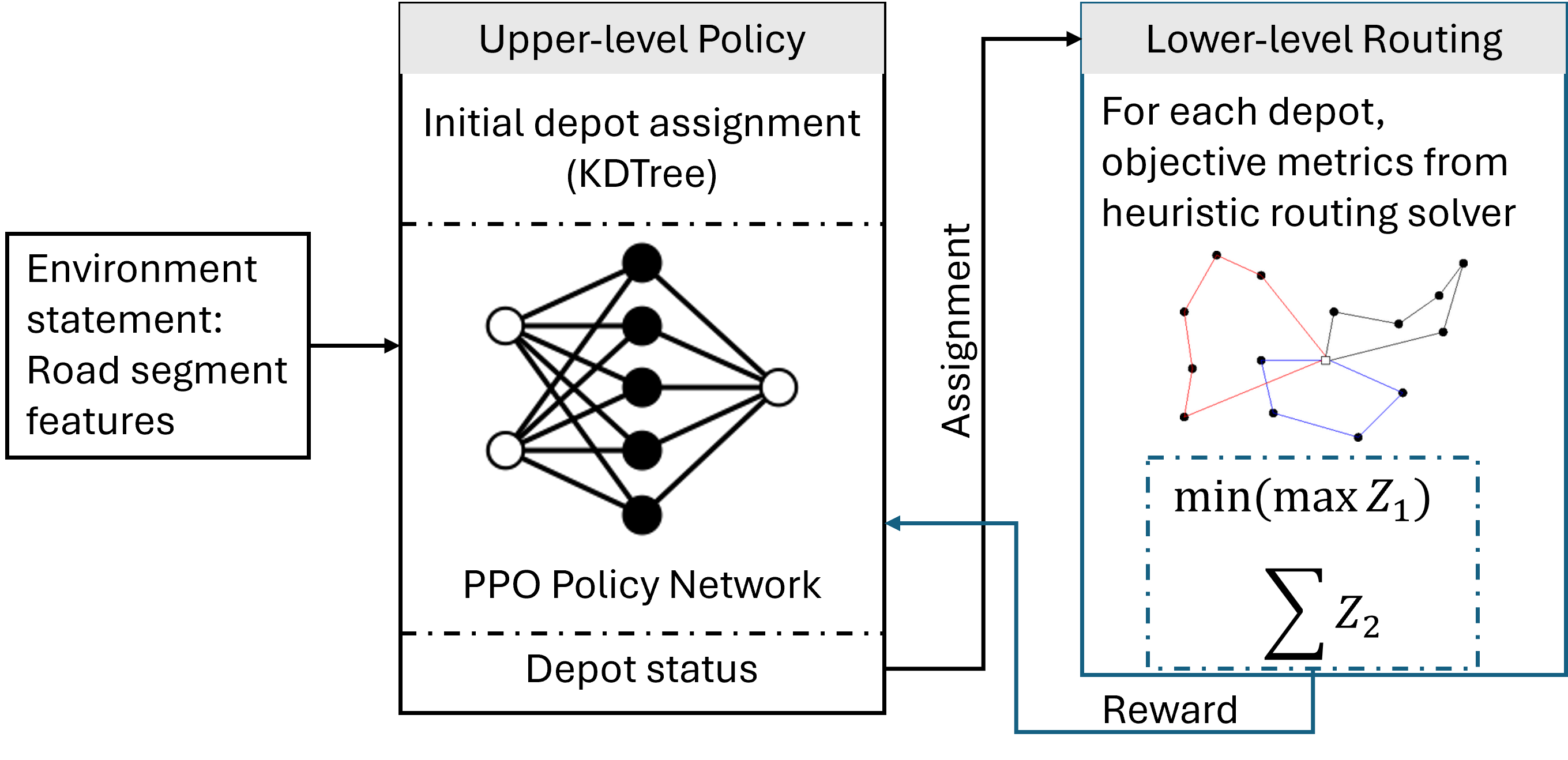}
    \caption{Overview of the bi-level optimization framework. The upper-level PPO agent assigns treatment-required road segments to depots. The lower-level routing module solves each subproblem under operational constraints and returns reward metrics (makespan and emissions) to guide policy updates}
    \label{fig:framework}
\end{figure}

All iterations persist artifacts for audit and A/B analysis: (i) edge-to-depot assignments, (ii) per-vehicle route CSVs (nodes/edges visited, distance, time, emissions), and (iii) an iteration log of $(Z1,Z2,r)$. Operators can override constraints (e.g., temporary depot closures or stock limits) via the same interface, and the pipeline ingests real network data (OSM-derived geometry, official depots, lane/speed attributes) used by the highway operator. Figure \ref{fig:framework} sketches this loop.

\section{Case Studies}

\subsection{Operation area} 

The case study covers a 60km x 60km area in England with coordinates--X from 444,000 to 504,000 and Y from 224,000 to 284,000--in British National Grid (EPSG:27700). It aims to demonstrate the proposed bi-level optimization approach with the strategic road network within this area, which is managed by National Highways. There are 3 service vehicle depots, Misterton depot, Pytchley depot, and Rothersthorpe depot within the study area. Interconnected local road networks are integrated in the graph model to allow vehicle turning and traversing.

\begin{figure}
    \centering
    \includegraphics[width=0.9\linewidth]{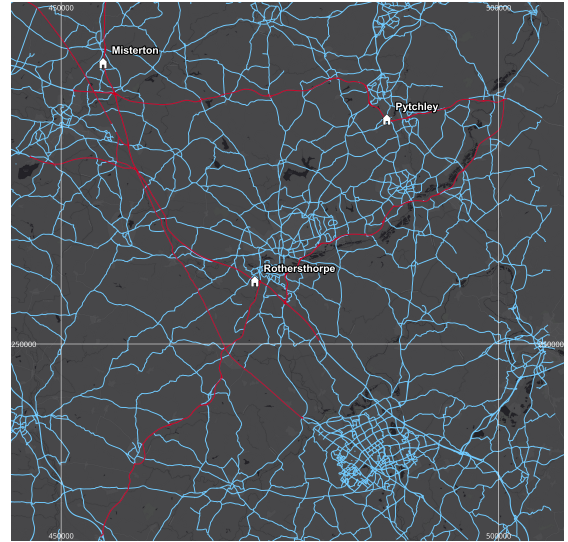}
    \caption{Operation area}
    \label{fig:enter-label}
\end{figure}

\subsection{Network Construction and Pre-processing}

We rebuild the network from OpenStreetMap (OSM) using GIS and a directed graph model, and identify the strategic road network edges that require treatment by map-matching historical National Highways' GPS traces to OSM links. Topological cleaning uses the largest strongly connected component to remove dangling artifacts and ensure depot reachability.

To reduce computational cost without altering routing semantics, we repeatedly merge maximal chains of degree-2 nodes whose adjacent edges share identical attributes (oneway, speed limit, lanes). Lengths are aggregated during merging. This compresses the graph from $99{,}536$ nodes and $176{,}636$ directed edges to $37{,}007$ nodes and $71{,}505$ directed edges, while preserving junction geometry and attributes needed for routing.

The resulting network is large and heterogeneous. It remains a single connected component and contains complete real-world junction topologies, including $207$ high-degree intersections (degree~$\ge 4$; max degree~6), reflecting the complex motorway structure. Edges span 5785km in total. Of these, $1515$ directed edges (543.5km) are marked for treatment, corresponding to 1208.7 lane-km after accounting for lane counts. Roughly $12.2\%$ of links are one-way. Speed limits cover seven discrete values from 10 to 70{mph} (median 50{mph}); lane counts are predominantly 1–2 lanes, with multi-lane motorway sections concentrated near interchanges. These factors, together with depot reachability and strict vehicle constraints (capacity, max distance, and 120{min} per-route duration), make single-level formulations computationally brittle at this scale.

\subsection{Results}

We evaluate two planners:
\begin{enumerate}
  \item \textbf{KDTree+NN}: static KDTree depot assignment followed by a nearest-neighbour (NN) routing heuristic; executed once without policy refinement.
  \item \textbf{KDTree-PPO+NN (10 iterations)}: identical first step (KDTree), then an iterative loop where a PPO policy adjusts assignment decisions using feedback from NN-produced routes. We run $10$ PPO–NN iterations, keeping all operational constraints unchanged.
\end{enumerate}

\begin{table}[th]
  \centering
  \caption{Raw results on the fixed network (no normalization).
  $Z_1$ = makespan (h); $Z_2$ = emissions (kg\,CO$_2$e).}
  \label{tab:results}
  \scalebox{0.95}{
  \begin{tabular}{lrrr}
    \hline
    Method & $Z_1$ (min) & $Z_2$ (kg CO$_2$) & NoV \\
    \hline
    KDTree+NN & 122.14 &  3,386.63 &  20\\
    KDTree-PPO+NN & 118.81 &  3,220.95 & 28 \\
    \hline
  \end{tabular}}
\end{table}

Since all experiments share the same network and fleet configuration:
(i) network coverage (treated road length in km and as \%), (ii) total distance (km), (iii) makespan (h), (iv) energy use (kWh) and emissions (kgCO\textsubscript{2}e) if available, (v) number of constraint violations (e.g., max shift, route duration), and (vi) computation time (min). We list the results of baseline approach and the results of the last iteration of our proposed approach in Table \ref{tab:results}.

Relative to the one-shot baseline, the developed bi-level optimization achieves simultaneous improvements in all metrics. These are Pareto improvements with respect to the reported objectives, indicating that PPO's iterative reassignment produces more coherent route clusters for the downstream NN-planer. The reduction in $Z_2$ (approximate $4\% - 5\%$) is consistent with less deadheading and idling, while the increase in NoV suggests better compliance with shift/route-duration constraints (lower overtime risk).

\begin{figure}
    \centering
    \includegraphics[width=0.95\linewidth]{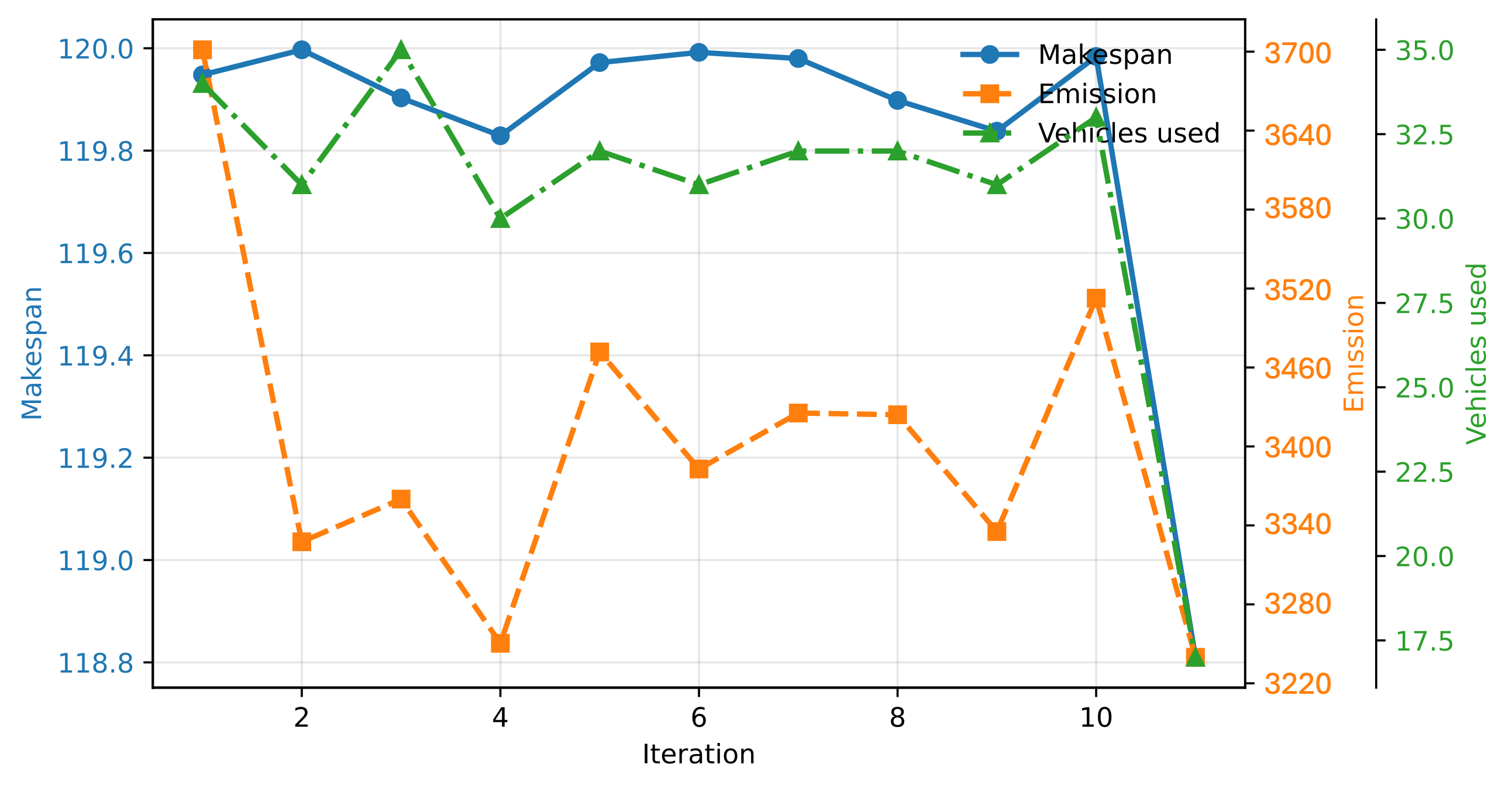}
    \caption{Objectives and fleet use across PPO–NN iterations.}
    \label{fig:coverage_curve}
\end{figure}

Although a $2\% - 3\%$ reduction in makespan ($Z1$) appears modest at the per-shift level, it compounds across vehicles and peak events, freeing capacity to treat residual hotspots and reducing depot congestion. The reduction of CO$_2$ per event is also meaningful in municipal reporting, especially when aggregated over a season. Because the NN-planner and all constraints are identical across methods, the gains can be attributed to PPO's assignment policy rather than differences in routing heuristics.

To examine how performance evolves within PPO, we track coverage after each PPO–NN interaction. Figure~\ref{fig:coverage_curve} shows the makespan (left y-axis), emissions(right y-axis), and vehicle used (offset right y-axis) over 11 iterations. Most improvement occurs early (around 1–4), with subsequent iterations refining assignments and stabilizing the metrics. This pattern indicates that PPO quickly captures the largest assignment gains and then fine-tunes depot boundaries, yielding shorter, more balanced routes with stable or reduced fleet utilization and lower operational burden overall.

\subsection{Route-level patterns and operational implications}

\begin{figure}[t!]
  \centering
  \begin{subfigure}[t]{0.225\textwidth}
    \centering
    \includegraphics[width=0.9\linewidth]{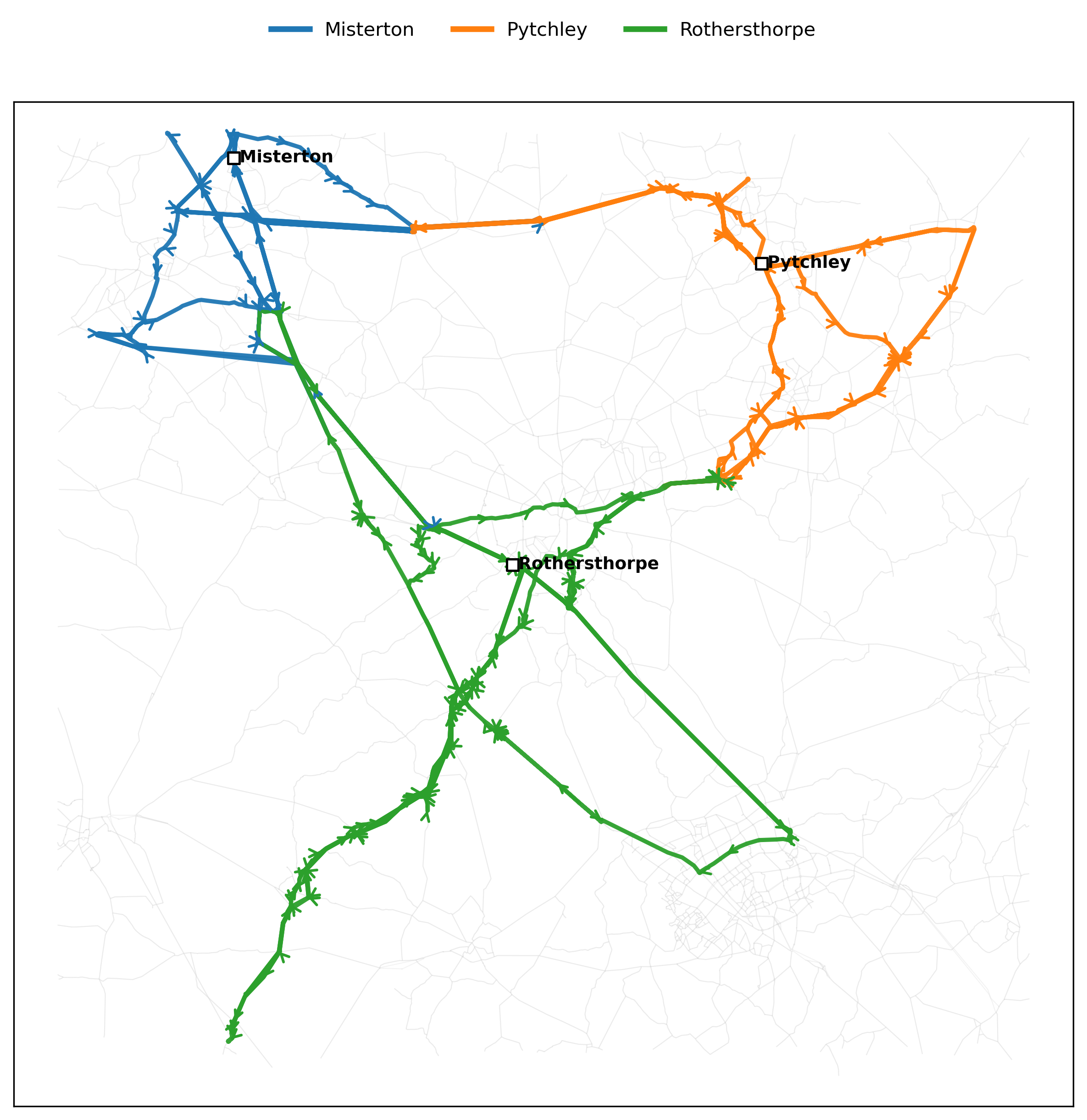}
    \caption{KDTree+NN (baseline).}
    \label{fig:kdtree_nn_routes}
  \end{subfigure}
  \begin{subfigure}[t]{0.225\textwidth}
    \centering
    \includegraphics[width=0.9\linewidth]{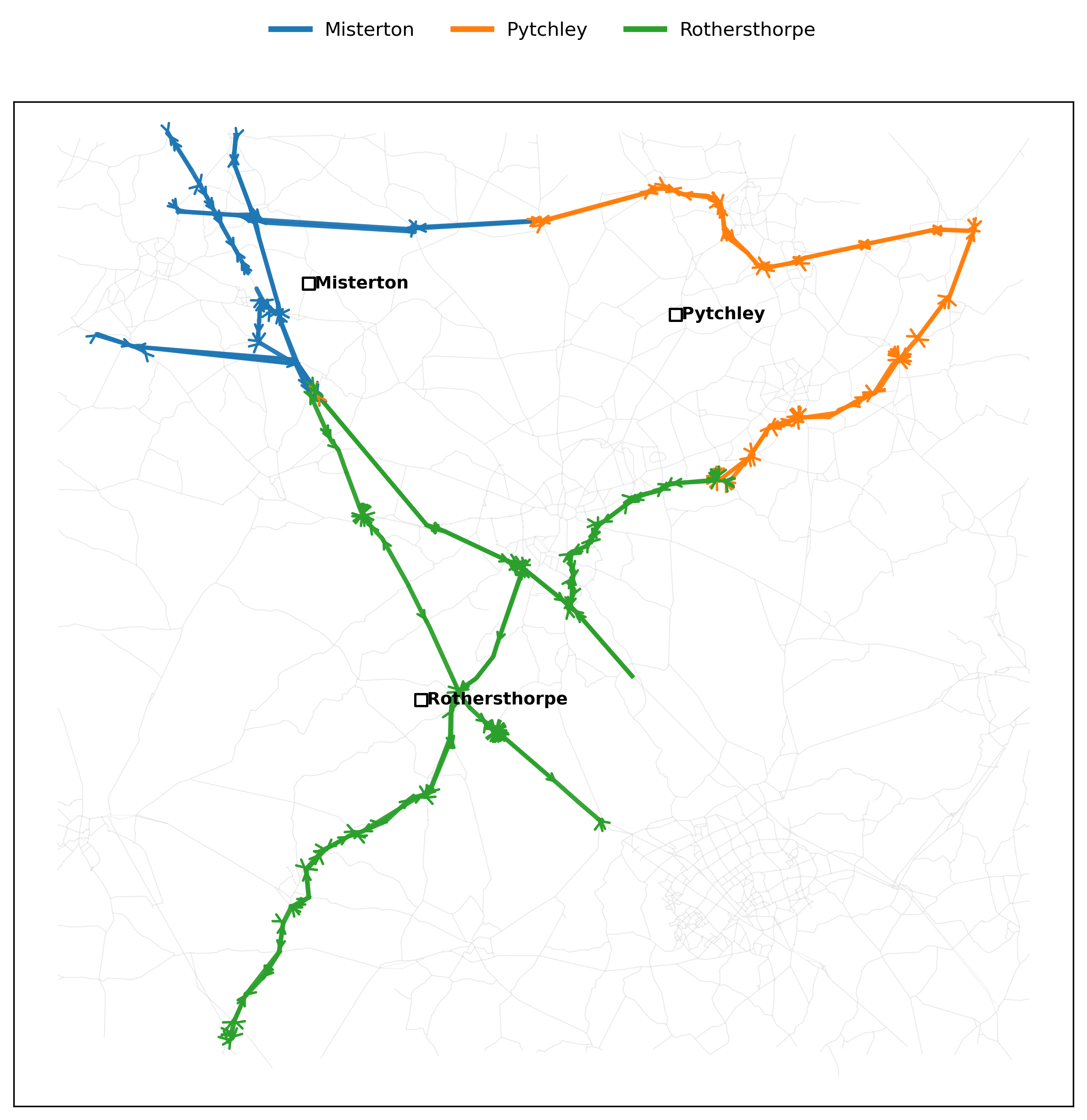}
    \caption{KDTree–PPO+NN after 10 iterations.}
    \label{fig:kdtree_ppo_nn_routes}
  \end{subfigure}
  \caption{Treated edges by depot. Coloured polylines show vehicle trajectories from each depot (blue = Misterton, orange = Pytchley, green = Rothersthorpe) drawn over the full road network (grey). Arrowheads indicate driving direction.}
  \label{fig:routes_side_by_side}
\end{figure}

Figure~\ref{fig:routes_side_by_side} contrasts the one-shot KDTree+NN baseline with the policy-refined KDTree–PPO+NN planner. The baseline exhibits long cross-depot incursions and several out-and-back stubs, particularly around the Rothersthorpe junction and along the eastern corridor. These artefacts create deadheading and coordination overhead across depots. After ten PPO–NN interactions, the depot territories become markedly more compact and depot-centric: Misterton’s coverage is consolidated in the northwest, Pytchley’s routes are confined to the eastern loop with fewer reversals, and Rothersthorpe’s vehicles service the south–central spokes with reduced overlap. The fewer hairpins and backtracks evident from the arrowheads align with the raw improvements reported earlier—shorter total distance and makespan, and stable or lower vehicle counts.

These maps make the application value concrete. First, the improvement is achieved without replacing existing dispatch logic: PPO only adjusts high-level assignment, while NN continues to produce feasible vehicle tours under the same field constraints. This preserves interpretability (clusters correspond to intuitive depot territories) and eases integration with current GIS and scheduling tools. Second, the gains occur in \emph{raw operational units} (treated kilometers, hours, vehicles), which directly translate to reduced overtime, lower fuel consumption, and fewer post-storm mop-up passes. Third, the iterative loop is computationally modest for pre-shift planning and can be warm-started from prior events, enabling routine deployment across weather cycles.

\section{Conclusion}

We presented a practical bi-level planner for winter road maintenance that layers a lightweight KDTree assignment, iterative PPO refinement, and an NN micro-router. On a 60$\times$60\,km UK strategic road network managed by our industry partner \emph{National Highways}, the approach consistently improved raw field metrics over a one-shot KDTree+NN baseline—shorter makespan and distance, comparable or lower fleet use—and produced depot-centric routes with fewer cross-boundary incursions. Crucially, deployment requires no change to existing dispatch practice: PPO adjusts only high-level assignments while the familiar NN tours remain feasible and auditable. The loop fits pre-shift planning, can be warm-started across events, and integrates cleanly with current GIS and scheduling tools.

Future work will fully leverage the scalability of bi-level approach for winter service optimization on a national level strategic road network, and explore incorporating storm progression and priority queues, depot stock/reload logistics, and limited mid-shift replanning—engineering additions to the same modular template—supporting wider roll-out across additional regions of the National Highways network.

\bibliography{iaai}

\end{document}